\documentclass{article}

 \PassOptionsToPackage{numbers, compress}{natbib}

 \usepackage[preprint]{nips_2018}

\usepackage{lipsum}
\usepackage[utf8]{inputenc} 
\usepackage[T1]{fontenc}    
\usepackage{hyperref}       
\usepackage{url}            
\usepackage{booktabs}       
\usepackage{amsfonts}       
\usepackage{nicefrac}       
\usepackage{microtype}      
\usepackage{multirow}
\usepackage{graphicx}
\usepackage{subcaption}
\usepackage{epstopdf}
\usepackage{multicol}

\title{Spatio-temporal Stacked LSTM for Temperature Prediction in Weather Forecasting}

\author{
  Zahra Karevan \\
  Department of Electrical Engineering\\
  KU Leuven\\
Leuven, Belgium
 \\
  \texttt{zahra.karevan@esat.kuleuven.be} \\
   \And
   Johan A. K. Suykens \\
	 Department of Electrical Engineering\\
   KU Leuven\\
    Leuven, Belgium
 \\
   \texttt{johan.suykens@esat.kuleuven.be} \\
	}

\begin{document}

\maketitle

\begin{abstract}
Long Short-Term Memory (LSTM) is a well-known method used widely on sequence learning and time series prediction. In this paper we deployed stacked LSTM model in an application of weather forecasting. We propose a 2-layer spatio-temporal stacked LSTM model which consists of independent LSTM models per location in the first LSTM layer. Subsequently, the input of the second LSTM layer is formed based on the combination of the hidden states of the first layer LSTM models. The experiments show that by utilizing the spatial information the prediction performance of the stacked LSTM model improves in most of the cases.

\end{abstract}

\section{Introduction}
The weather system is a challenging complex system and state-of-the-art methods utilize Numerical Weather Prediction (NWP) which is a computationally intense method \cite{bauer2015quiet}. Recently, data-driven approaches for weather forecasting have become a major interest of researchers as they are computationally simpler and more straightforward \cite{feng2017data,Houthuys2017,hu2014pattern,ZahraESANN2017}. 

Long Short-Term Memory (LSTM), proposed by Hochreiter $\&$ Schmidhuber \cite{hochreiter1997long}, is a popular type of recurrent neural network which is able to capture long-term dependencies. LSTMs have been widely used and shown significant performance on different sequence learning problems and time series prediction \cite{ng2015beyond,sutskever2014sequence, graves2013speech,freeman2018forecasting,lipton2015learning, tian2015predicting}. LSTMs have been also used in analyzing spatio-temporal datasets \cite{patraucean2015spatio, liu2016spatio}. Stacked LSTM is a deep architecture which consists of more than one layer of LSTM and the input of each LSTM layer is the hidden states of the previous LSTM layer \cite{graves2013hybrid,sutskever2014sequence}. 

In this paper a spatio-temporal stacked LSTM model is proposed and its performance is evaluated on the application of temperature prediction. In the proposed model independent LSTM models per location are trained and afterward, the input of the second layer LSTM is formed based on the combination of the hidden states of the LSTM models in the first layer. It is worth mentioning that the general structure of the spatio-temporal stacked LSTM is similar to the approach proposed by Su et al. in the framework of Convolutional Neural Networks for 3D shape recognition \cite{su2015multi}. Note that both stacked LSTM and spatio-temporal stacked LSTM methodologies can be explained as a multi-view approach as, similar to \cite{Houthuys2017}, they are fusing the information of different cities. Stacked LSTM and spatio-temporal stacked LSTM benefit from early fusion and intermediate fusion of the information from different views respectively.

\section{Spatio-temporal Stacked LSTM}
Assuming $i_{t}^{[l]}$, $f_{t}^{[l]}$, $o_{t}^{[l]}$, $c_{t}^{[l]}$ and $h_{t}^{[l]}$ to be the values of the input gate, forget gate, output gate, memory cell and hidden state at time ${t}$ in the sequence and layer $l$ respectively, and $x_{t,k}$ be the input of the system at time ${t}$ at location $k$, the stacked LSTM model based on the architecture of the LSTM cell defined in \cite{graves2013generating} is shown in Table \ref{StackedLSTMEq}. Note that $x_t$ as an input of the model is a concatenation of the variables from all locations; i.e. $x_t = [x_{t,1},x_{t,2},\cdots,x_{t,c}]^T $.
In this study we focus on a 2-layer stacked LSTM; however, the methodology can be extended to a larger number of layers. The full weight matrices $W_{xj}$ for $j \in \{i,f,o,c\}$ are the weights that connect the input to the corresponding gates and the memory cell. The weight matrices $W_{cj}$ for $j \in \{i,f,o\}$ are diagonal matrices that connect the cell memory to different gates. Note that the number of neurons for all gates is a predefined parameter and the equations are applied for each neuron. 
For simplicity, we use column vectors $w_{\rm lstm}^{[l]}$ and $b_{\rm lstm}^{[l]}$ to indicate all the elements in $\{W_{xi}^{[l]}, W_{ci}^{[l]}, W_{xf}^{[l]}, W_{hf}^{[l]}, W_{cf}^{[l]}, W_{xc}^{[l]}, W_{hc}^{[l]}, W_{xo}^{[l]}, W_{ho}^{[l]}, W_{co}^{[l]}\}$ and $\{b_i^{[l]},b_f^{[l]},b_c^{[l]},b_o^{[l]}\}$ respectively.
\begin{table}[h]
  \caption{Equations of the stacked LSTM}
	\label{StackedLSTMEq}
\resizebox{\columnwidth}{!}{
\begin{tabular}{p{0.5cm}|p{8cm}|p{8cm}|}
\cline{2-3}
                                        &        Layer 1 LSTM          &         Layer 2 LSTM          \\ \hline
\multicolumn{1}{|l|}{Input}                  &   $x_t = [x_{t,1},x_{t,2},\cdots,x_{t,c}] $             &        $h_t ^{[1]} $          \\ \hline
\multicolumn{1}{|l|}{\multirow{5}{*}{Equations}} &$i_{t}^{[1]} = \sigma(W_{xi}^{[1]}x_{t} + W_{hi}^{[1]}h_{{t}-1}^{[1]} + W_{ci}^{[1]}c_{{t}-1}^{[1]} +b_i^{[1]})$ & $i_{t}^{[2]} = \sigma(W_{h^{[1]}i}^{[2]}h_{t}^{[1]} + W_{hi}^{[2]}h_{{t}-1}^{[2]} + W_{ci}^{[2]}c_{{t}-1}^{[2]} +b_i^{[2]})$ \\
\multicolumn{1}{|l|}{}                  &  $f_{t}^{[1]} = \sigma(W_{xf}^{[1]}x_{t} + W_{hf}^{[1]}h_{{t}-1}^{[1]} + W_{cf}^{[1]}c_{{t}-1}^{[1]} +b_f^{[1]})$   &       $f_{t}^{[2]} = \sigma(W_{xf}^{[2]}h_{t}^{[1]} + W_{hf}^{[2]}h_{{t}-1}^{[2]} + W_{cf}^{[2]}c_{{t}-1}^{[2]} +b_f^{[2]})$            \\
\multicolumn{1}{|l|}{}                  &   $c_{t}^{[1]} = f_{t}^{[1]} \odot c_{t-1}^{[1]}+i_{t}^{[1]} \odot \tanh(W_{xc}^{[1]}x_{t}+ W_{hc}^{[1]}h_{{t}-1}^{[1]}+b_c^{[1]})$    & $c_{t}^{[2]} = f_{t}^{[2]} \odot c_{t-1}^{[2]}+i_{t}^{[2]} \odot \tanh(W_{xc}^{[2]}h_{t}^{[1]}+ W_{hc}^{[2]}h_{{t}-1}^{[2]}+b_c^{[2]})$                    \\
\multicolumn{1}{|l|}{}                  & $o_{t}^{[1]} = \sigma(W_{xo}^{[1]}x_{t} + W_{ho}^{[1]}h_{{t}-1}^{[1]} + W_{co}^{[1]}c_{{t}}^{[1]} +b_o^{[1]})$ &     $o_{t}^{[2]} = \sigma(W_{xo}^{[2]}h_{t}^{[2]} + W_{ho}^{[2]}h_{{t}-1}^{[2]} + W_{co}^{[2]}c_{{t}}^{[2]} +b_o^{[2]})$               \\ 
\multicolumn{1}{|l|}{}                  & $h_{t}^{[1]} = o_{t}^{[1]} \odot \tanh(c_{t}^{[1]})$  & $h_{t}^{[2]} = o_{t}^{[2]} \odot \tanh(c_{t}^{[2]})$ \\ \hline
\multicolumn{1}{|l|}{\multirow{2}{*}{Summary}} & $c_t^{[1]} = f(c_{t-1}^{[1]},h_{t-1}^{[1]}, x_t; w_{\rm lstm}^{[1]},b_{\rm lstm}^{[1]}) $ &$c_t^{[2]} = f(c_{t-1}^{[2]},h_{t-1}^{[2]}, h_t^{[1]}; w_{\rm lstm}^{[2]},b_{\rm lstm}^{[2]}) $\\ 
\multicolumn{1}{|l|}{}   &   $ h_{{t}}^{[1]} = g(h_{{t}-1}^{[1]},c_{{t}-1}^{[1]},x_{{t}}; w_{\rm lstm}^{[1]},b_{\rm lstm}^{[1]})$ &  $h_{{t}}^{[2]} = g(h_{{t}-1}^{[2]},c_{{t}-1}^{[2]},h_{{t}}^{[1]}; w_{\rm lstm}^{[2]},b_{\rm lstm}^{[2]})$  \\ \hline
\end{tabular}}
\end{table}

\begin{figure}[]
\centering
\begin{subfigure}{.3\textwidth}
  \centering
	\vspace{0.3cm}
  \includegraphics[width=4.5cm,height=3.5cm]{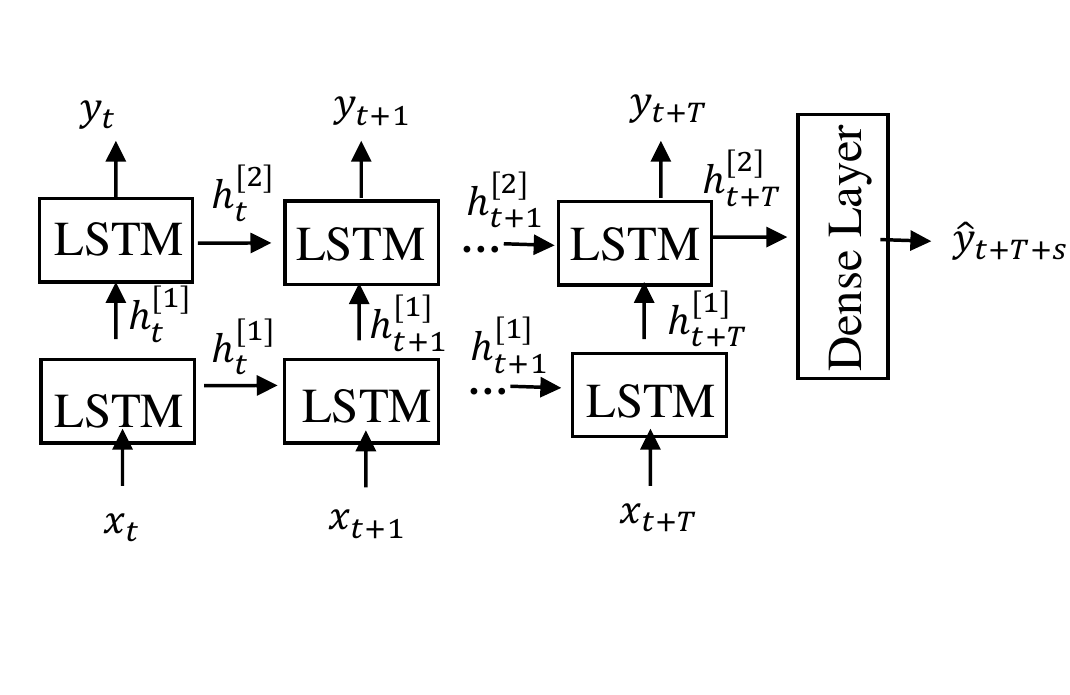}
	\vspace{0.3cm}
  \caption{Two-layer stacked LSTM}
  \label{fig:sfig1}
\end{subfigure}%
\centering
\begin{subfigure}{.6\textwidth}
\centering
  \includegraphics[width=8cm,height=4.5cm]{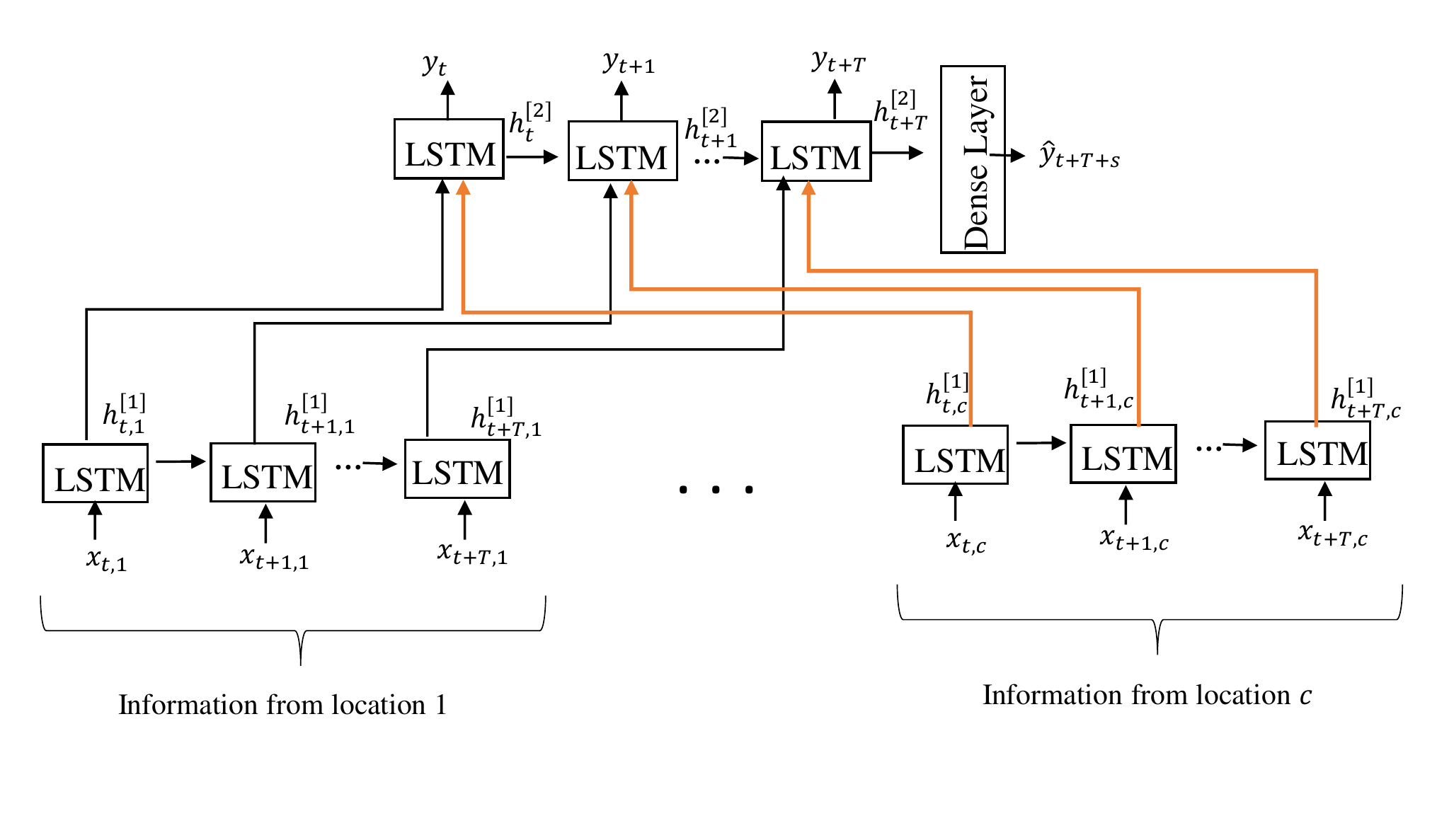}
  \caption{Two-layer spatio-temporal stacked LSTM}
  \label{fig:sfig2}
\end{subfigure}
\caption{The scheme of the stacked LSTM and the proposed spatio-temporal stacked LSTM models when the number of layers is equal to two.}
\label{fig:stackvsST}
\end{figure}

In Table \ref{STStackedLSTMEq} the equations for the proposed spatio-temporal stacked LSTM model are shown. As is explained, instead of one LSTM model in the first layer, there are independent LSTM models per location. Hence, having 5 locations, 5 LSTM models are created. The full weight matrices $W_{xj,k}$ for $j \in \{i,f,o,c\}$ refer to the connection of the data of location $k$ to different gates in the corresponding LSTM model. Similarly, other weight matrices, biases, gate values, memory cell and hidden state used subscript $k$ to indicate that they correspond to the LSTM related to the location $k$. For simplicity, we use column vectors $w_{{\rm lstm},k}^{[l]}$ and $b_{{\rm lstm},k}^{[l]}$ which include all the elements in $\{W_{xi,k}^{[l]}, W_{ci,k}^{[l]}, W_{xf,k}^{[l]}, W_{hf,k}^{[l]}, W_{cf,k}^{[l]}, W_{xc,k}^{[l]}, W_{hc,k}^{[l]}, W_{xo,k}^{[l]}, W_{ho,k}^{[l]}, W_{co,k}^{[l]}\}$ and $\{b_{i,k}^{[l]},b_{f,k}^{[l]},b_{c,k}^{[l]},b_{o,k}^{[l]}\}$ to refer to the parameters for the LSTM part related to location $k$ for $k \in \{1,2,\cdots c\}$ in the first layer. The information from different locations are then combined by merging the hidden states of the first layer and passing it as input to the second layer. For the second LSTM layer the definition of $w_{\rm lstm}^{[l]}$ and $b_{\rm lstm}^{[l]}$ remains the same. Figure \ref{fig:stackvsST} depicts the stacked LSTM and spatio-temporal stacked model when the number of layers is equal to two. As is shown, for the stacked LSTM model, the hidden states of the first layer are used as the input of the second layer. For the second LSTM layer the definition of $w_{\rm lstm}^{[l]}$ and $b_{\rm lstm}^{[l]}$ remains the same. However, in case of spatio-temporal stacked LSTM, there are independent LSTM models per location, and afterward the input of the second LSTM layer is defined based on the combination of the hidden states of the LSTM models of the first layer. Note that if the number of layers is more that two, the merging of the hidden states is possible at any layer before the last LSTM layer; e.g. if we have a 3-layer spatio-temporal stacked LSTM model, the combination of the hidden states can happen after the first LSTM layer or the second one. Note that in both stacked LSTM and spatio-temporal stacked LSTM models, after the second LSTM layer, a dense layer is used. The final prediction can be done by using $\hat{y}_{t+T+q} = w_{\rm dense}^Th_{t+T}^{[2]}+b_{\rm dense}^{[2]}$ where $q$ is the number of days ahead to predict, $T$ is the input sequence length, and $w_{\rm dense}$ and $b_{\rm dense}$ are the weights and bias term in the dense layer. For the experiments, we use a quadratic loss function to train the network and utilize $L2$-norm regularization to avoid overfitting.

\begin{table}[h]
\caption{Equations of the spatio-temporal stacked LSTM}
	\label{STStackedLSTMEq}
\resizebox{\columnwidth}{!}{
\begin{tabular}{p{0.5cm}|p{8cm}|p{8cm}|}
\cline{2-3}
                                        &        Layer 1 LSTM (for $k \in \{1,2, \cdots, c\}$ )          &         Layer 2 LSTM          \\ \hline
\multicolumn{1}{|l|}{Input}                  &   $x_{t,k} $            &        $h_t ^{[1]} = [h_{t,1} ^{[1]},h_{t,2} ^{[1]},\cdots,h_{t,c} ^{[1]}]  $          \\ \hline
\multicolumn{1}{|l|}{\multirow{5}{*}{Equations}} &$i_{t,k}^{[1]} = \sigma(W_{xi,k}^{[1]}x_{t,k} + W_{hi,k}^{[1]}h_{{t}-1,k}^{[1]} + W_{ci,k}^{[1]}c_{{t}-1,k}^{[1]} +b_{i,k}^{[1]})$ & $i_{t}^{[2]} = \sigma(W_{h^{[1]}i}^{[2]}h_{t}^{[1]} + W_{hi}^{[2]}h_{{t}-1}^{[2]} + W_{ci}^{[2]}c_{{t}-1}^{[2]} +b_{i}^{[2]})$ \\
\multicolumn{1}{|l|}{}                  &  $f_{t,k}^{[1]} = \sigma(W_{xf,k}^{[1]}x_{t,k} + W_{hf,k}^{[1]}h_{{t}-1,k}^{[1]} + W_{cf,k}^{[1]}c_{{t}-1,k}^{[1]} +b_{f,k}^{[1]})$   &       $f_{t}^{[2]} = \sigma(W_{xf}^{[2]}h_{t}^{[1]} + W_{hf}^{[2]}h_{{t}-1}^{[2]} + W_{cf}^{[2]}c_{{t}-1}^{[2]} +b_f^{[2]})$            \\
\multicolumn{1}{|l|}{}                  &   $c_{t,k}^{[1]} = f_{t,k}^{[1]} \odot c_{t-1,k}^{[1]}+i_{t,k}^{[1]} \odot \tanh(W_{xc,k}^{[1]}x_{t,k}+ W_{hc,k}^{[1]}h_{{t}-1,k}^{[1]}+b_{c,k}^{[1]})$    & $c_{t}^{[2]} = f_{t}^{[2]} \odot c_{t-1}^{[2]}+i_{t}^{[2]} \odot \tanh(W_{xc}^{[2]}h_{t}^{[1]}+ W_{hc}^{[2]}h_{{t}-1}^{[2]}+b_c^{[2]})$                    \\
\multicolumn{1}{|l|}{}                  & $o_{t,k}^{[1]} = \sigma(W_{xo,k}^{[1]}x_{t,k} + W_{ho,k}^{[1]}h_{{t}-1,k}^{[1]} + W_{co,k}^{[1]}c_{{t},k}^{[1]} +b_{o,k}^{[1]})$ &     $o_{t}^{[2]} = \sigma(W_{xo}^{[2]}h_{t}^{[2]} + W_{ho}^{[2]}h_{{t}-1}^{[2]} + W_{co}^{[2]}c_{{t}}^{[2]} +b_o^{[2]})$               \\ 
\multicolumn{1}{|l|}{}                  & $h_{t,k}^{[1]} = o_{t,k}^{[1]} \odot \tanh(c_{t,k}^{[1]})$  & $h_{t}^{[2]} = o_{t}^{[2]} \odot \tanh(c_{t}^{[2]})$ \\ \hline
\multicolumn{1}{|l|}{\multirow{2}{*}{Summary}} & $c_{t,k}^{[1]} = f(c_{t-1,k}^{[1]},h_{t-1,k}^{[1]}, x_{t,k}; w_{{\rm lstm},{k}}^{[1]},b_{{\rm lstm},{k}}^{[1]}) $ &$c_t^{[2]} = f(c_{t-1}^{[2]},h_{t-1}^{[2]}, h_t^{[1]}; w_{\rm lstm}^{[2]},b_{\rm lstm}^{[2]}) $\\ 
\multicolumn{1}{|l|}{}   &   $ h_{{t},k}^{[1]} = g(h_{{t}-1,k}^{[1]},c_{{t}-1,k}^{[1]},x_{{t,k}}; w_{{\rm lstm},{k}}^{[1]},b_{{\rm lstm},{k}}^{[1]})$ &  $h_{{t}}^{[2]} = g(h_{{t}-1}^{[2]},c_{{t}-1}^{[2]},h_{{t}}^{[1]}; w_{\rm lstm}^{[2]},b_{\rm lstm}^{[2]})$  \\ \hline
\end{tabular}}
\end{table}

One of the advantages of the proposed spatio-temporal stacked LSTM method is a smaller number of parameters in comparison with stacked LSTM. Assume the total number of neurons in the first layer and second layer is similar in both cases; in other words, if the number of neurons in stacked LSTM model in the first layer is $n_1$, then in the spatio-temporal stacked LSTM, each LSTM model in the first layer has $\frac{n_1}{c}$ neurons where $c$ is the number of locations. In this case, the spatio-temporal stacked LSTM is similar to the case that the full weight matrices in stacked LSTM are considered to be block diagonal at which point each block is related to a location. Hence, the number of parameters to be optimized is smaller in the proposed method. This makes the spatio-temporal stacked LSTM a better choice when the number of samples in the training set is relatively small. On the other hand, in spatio-temporal stacked LSTM, the relation between the locations are taken into account in the second LSTM layer by combining the hidden states from the first layer.  


%
%

\section{Experiments}

\begin{multicols}{2}
In this paper, the data have been collected from the Weather Underground company website \cite{Wunderground} and cover a time period from the beginning of 2007 to mid-2014 for 5 cities including Brussels, Antwerp, Liege, Amsterdam and Eindhoven. 
To evaluate the performance of the proposed methods in various weather conditions, two test sets are defined: (i) from mid-November 2013 to mid-December 2013 ({Nov/Dec}) and (ii) from mid-April 2014 to mid-May 2014 ({Apr/May}).
The data contain 18 measured weather variables, such as temperature and humidity, for each day per city. In order to benefit from all available data, the training data that is used for each test set includes the data from the beginning of 2007 until the day before the corresponding test set.

\centering
		\includegraphics[width=5cm,height=4.5cm]{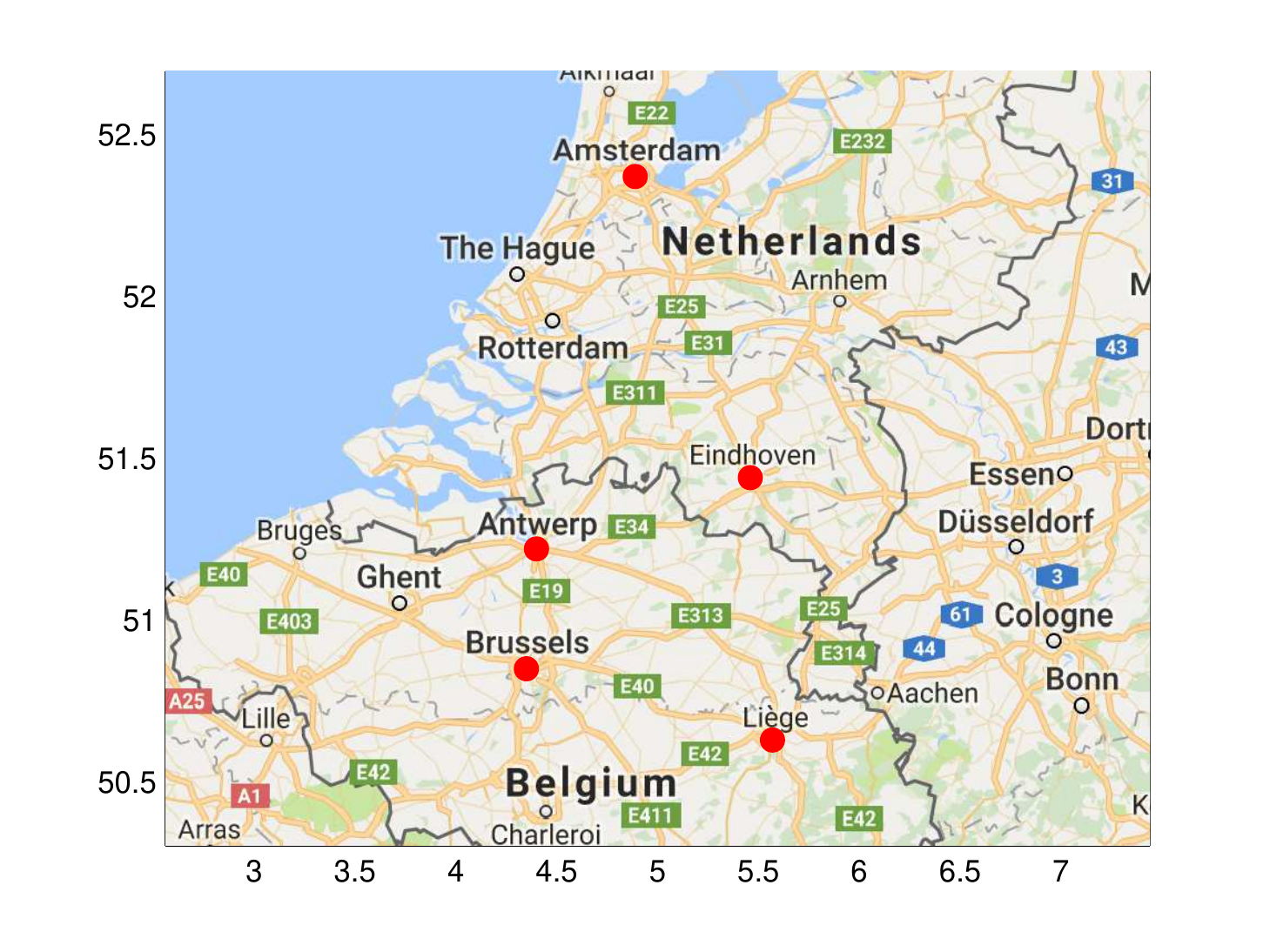}
		\captionof{figure}{Weather stations (picture: Google maps) \label{stations}}

\end{multicols}


In this study, the LSTM cell architecture described by Zaremba $\&$ Vinyals \cite{zaremba2014recurrent} implemented in TensorFlow has been used for the experiments. The considered range for the tuning parameters were selected empirically. For the number of neurons, in the stacked LSTM we examined $\{20,40,80,160,320,640\}$ in the first layer and $\{ 32,64,128,256\}$ for the second layer. In case of the spatio-temporal stacked LSTM, the number of neurons in LSTM per location in the first layer is considered to be in the set of $\{4,8,16,32,64,128\}$. Note that as there are 5 locations in the first layer, the total number of neurons in the first layer are similar in both models. For the inner state, we deployed both tanh and sigmoid as the activation function. In the experiments, the sequence length is considered to be 10.

The experiments were conducted for the prediction of the minimum and maximum temperature in Brussels for 1 to 6 days ahead. To avoid local minima problems in neural networks, the experiments are repeated 5 times and the median Mean Absolute Error (MAE) and the Mean Squared Error (MSE) on both test sets are presented in Table \ref{NovDec}. As is shown, using sigmoid as the inner activation function can result in better performance. Moreover, it can be seen that in most of the cases taking the spatial information into account can improve the performance. This is more evident in case the activation function in the inner state is tanh.

\begin{table}
  \caption{MAE of the stacked LSTM and the Spatio-Temporal stacked LSTM (ST stacked LSTM) for min. and max. temperature prediction in Nov/Dec and Apr/May test sets.}
  \label{NovDec}
  \centering
	\resizebox{\columnwidth}{!}{
 \begin{tabular}{p{1cm}p{0.7cm}p{0.7cm}|p{2cm}|p{2cm}|p{2cm}|p{2cm}|p{2cm}|p{2cm}|p{2cm}|p{2cm}|}
	\cline{4-11}
                               &         &  & \multicolumn{4}{p{8cm}|}{Activation function :tanh}    & \multicolumn{4}{p{8cm}|}{Activation function : sigmoid}      \\ \cline{4-11} 
                                &        &  & \multicolumn{2}{p{4cm}|}{MAE} & \multicolumn{2}{p{4cm}|}{MSE} & \multicolumn{2}{p{4cm}|}{MAE} & \multicolumn{2}{p{4cm}|}{MSE} \\ \hline
\multicolumn{1}{|p{0.7cm}|}{Testset}&{Steps ahead}       & Temp. &    stacked LSTM     &    ST stacked LSTM&stacked LSTM     &    ST stacked LSTM &stacked LSTM     &    ST stacked LSTM & stacked LSTM     & ST stacked LSTM \\ \hline
\multicolumn{1}{|c|}{\multirow{12}{*}{Nov/Dec}}&{\multirow{2}{*}{1}} & Min &   1.66       &  \textbf{1.43}         &  4.36         &    \textbf{ 3.64}      &      1.69     &     \textbf{ 1.56}     &     4.04      &   \textbf{3.71}        \\ \cline{3-11} 
\multicolumn{1}{|c|}{}        &          &  Max&   \textbf{1.15}        &    1.22       &    \textbf{2.48}       &    2.65       &        1.37   &      \textbf{ 1.23}    &     3.87      &    \textbf{ 2.96}      \\ \cline{2-11}
\multicolumn{1}{|c|}{} & {\multirow{2}{*}{2}} & Min &    2.30       &          \textbf{1.72} &   8.17        &    \textbf{4.36}       &  1.86         &  \textbf{1.76}         &     5.12      &      \textbf{5.00}     \\ \cline{3-11} 
\multicolumn{1}{|c|}{} &                 & Max &  1.89         &    \textbf{1.71}       &  6.51         &   \textbf{4.57}        &  1.65         &  \textbf{1.61 }        &    \textbf{4.24}       &   4.99        \\ \cline{2-11}
\multicolumn{1}{|c|}{}&{\multirow{2}{*}{3}} & Min &  {3.04 }        &    \textbf{1.72}       & {14.16}          &     \textbf{4.22}      &    \textbf{1.94 }      &        \textbf{1.94}   &  5.35         &   \textbf{5.23}        \\ \cline{3-11} 
\multicolumn{1}{|c|}{} &                 & Max &  3.44         &  \textbf{1.86}        &  17.30         &    \textbf{4.83 }      &    \textbf{1.73  }     &      \textbf{1.73 }    &   \textbf{4.99}        &  5.34         \\ \cline{2-11}
\multicolumn{1}{|c|}{}&{\multirow{2}{*}{4}} & Min &   3.28        & \textbf{1.98}          &  17.52         &   \textbf{6.39}        &    1.66       &   \textbf{1.57}        &   4.16        & \textbf{3.74}          \\ \cline{3-11} 
\multicolumn{1}{|c|}{} &                 & Max &  2.56         &   \textbf{2.14}        &   9.36        &   \textbf{5.87}        &  \textbf{ 1.61}        &    1.76       &    \textbf{ 3.85}      &   3.87        \\ \cline{2-11}
\multicolumn{1}{|c|}{}&{\multirow{2}{*}{5}} &Min &   3.72        &  \textbf{ 1.71 }       &   22.20        &      \textbf{4.39}     &     \textbf{1.58}      &  \textbf{ 1.58}        &    \textbf{4.06 }      &     4.13      \\ \cline{3-11} 
\multicolumn{1}{|c|}{} &                 & Max &     2.75      &  \textbf{1.89}         &   11.30        &      \textbf{4.92 }    &   1.58        &     \textbf{1.55}      &      \textbf{ 3.51}    &3.65           \\ \cline{2-11}
\multicolumn{1}{|c|}{}&{\multirow{2}{*}{6}}  &Min &    3.23       &  \textbf{ 1.90}        &    14.06       &  \textbf{5.42}         &\textbf{1.76}           &   1.90        &    \textbf{4.68}       &  5.09         \\ \cline{3-11} 
\multicolumn{1}{|c|}{}      &            & Max &   4.01        &   \textbf{1.80}        &  22.06         &     \textbf{5.40}      &  \textbf{1.63}         &    1.68       &   \textbf{4.70}        &   4.79        \\ \hline \hline

\multicolumn{1}{|c|}{\multirow{12}{*}{Apr/May}}&{\multirow{2}{*}{1}} & Min &  1.64         &   \textbf{1.60}        &   \textbf{4.15}        &     4.63      &   1.58        &   \textbf{1.55}        &   \textbf{3.78 }       &    3.92       \\ \cline{3-11} 
\multicolumn{1}{|c|}{}                  & & Max&  \textbf{ 2.27  }      &    2.45       &\textbf{7.66}           &     8.51      & \textbf{  2.24}        &   2.27        &  8.00         & \textbf{ 7.93  }       \\ \cline{2-11}
\multicolumn{1}{|c|}{} & {\multirow{2}{*}{2}} & Min & 2.52          &   \textbf{2.15 }       &  12.36         &     \textbf{9.09}      &   2.01        &   \textbf{1.96   }     &  7.86         &  \textbf{ 7.75 }       \\ \cline{3-11} 
\multicolumn{1}{|c|}{}                  && Max &  \textbf{2.64 }        &   \textbf{2.64 }       & 11.87          &    \textbf{10.37}       &    2.77       &  \textbf{ 2.55  }      & 10.03          &   \textbf{ 9.38 }      \\ \cline{2-11}
\multicolumn{1}{|c|}{} & {\multirow{2}{*}{3}} & Min &   2.83        &  \textbf{2.20 }        &    14.58       &      \textbf{9.07}     &     2.09      &\textbf{2.03}& 8.86          &  \textbf{ 8.54 }       \\ \cline{3-11} 
\multicolumn{1}{|c|}{}                  && Max &  3.61         &   \textbf{3.03  }      & 18.63          &     \textbf{12.75}      &   \textbf{2.53 }       &   2.58        &    \textbf{8.98 }      &    9.29       \\ \cline{2-11}
\multicolumn{1}{|c|}{} & {\multirow{2}{*}{4}} & Min &  2.63         & \textbf{ 2.25 }        &    12.09       &    \textbf{ 9.59}      &    \textbf{2.03}       &2.07           &  \textbf{8.27  }       &   8.36        \\ \cline{3-11} 
\multicolumn{1}{|c|}{}                  && Max &  3.08         &    \textbf{ 2.92}      &  13.02         &   \textbf{12.69}        &  2.72         &   \textbf{ 2.59}       &  10.18         &  \textbf{  9.75 }      \\ \cline{2-11}
\multicolumn{1}{|c|}{} & {\multirow{2}{*}{5}} & Min &   2.63        &   \textbf{2.51}        &    \textbf{11.71}       &      12.36     &     2.36      &       \textbf{2.31}    &     10.17      &  \textbf{10.05 }        \\ \cline{3-11} 
\multicolumn{1}{|c|}{}                 & & Max &  3.62         &    \textbf{ 2.89  }    &     19.82      &    \textbf{12.89}       &    \textbf{2.64 }      &      2.99     &      \textbf{10.17}     &   13.46        \\ \cline{2-11}
\multicolumn{1}{|c|}{} & {\multirow{2}{*}{6}} & Min & 3.04          &   \textbf{2.93  }      &    15.58       &     \textbf{15.39}      &  \textbf{2.46}         &    2.53       &    \textbf{ 10.92}      &   11.99        \\ \cline{3-11} 
\multicolumn{1}{|c|}{}                  && Max & \textbf{2.77   }       &    2.99       &   \textbf{10.99}        &      13.51     &     3.04      &   \textbf{2.95}        & 12.58          &    \textbf{ 12.15 }     \\ \hline

\end{tabular}}
\end{table}

 In addition, the comparison between the performance of the stacked LSTM, the proposed method and Weather Underground over the two test sets together is depicted in Figure \ref{ResAllST}. 
  \begin{multicols}{2}
      As is shown, although very few locations are taken into account, the LSTM models (as data-driven approaches) outperform the state-of-the-art method used for minimum temperature prediction. Also, for maximum temperature prediction the performance of the LSTM models are competitive with the performance of the state-of-the-art methods. 
		\vspace{-0.5cm}
		\section{Conclusion}
				\vspace{-0.4cm}
In this study, we proposed a spatio-temporal stacked LSTM model at which point in the first layer different LSTM models are considered per location, and then the corresponding hidden states are merged and given as input to the next layer. The proposed method was deployed in an application of weather forecasting.
 
The experimental results suggest that considering spatio-temporal property of the data in the LSTM model can improve the performance of the prediction. Moreover, it is shown that the proposed method is competitive with the state-of-the-art method in weather forecasting.

		\includegraphics[width=7.5cm,height=6cm]{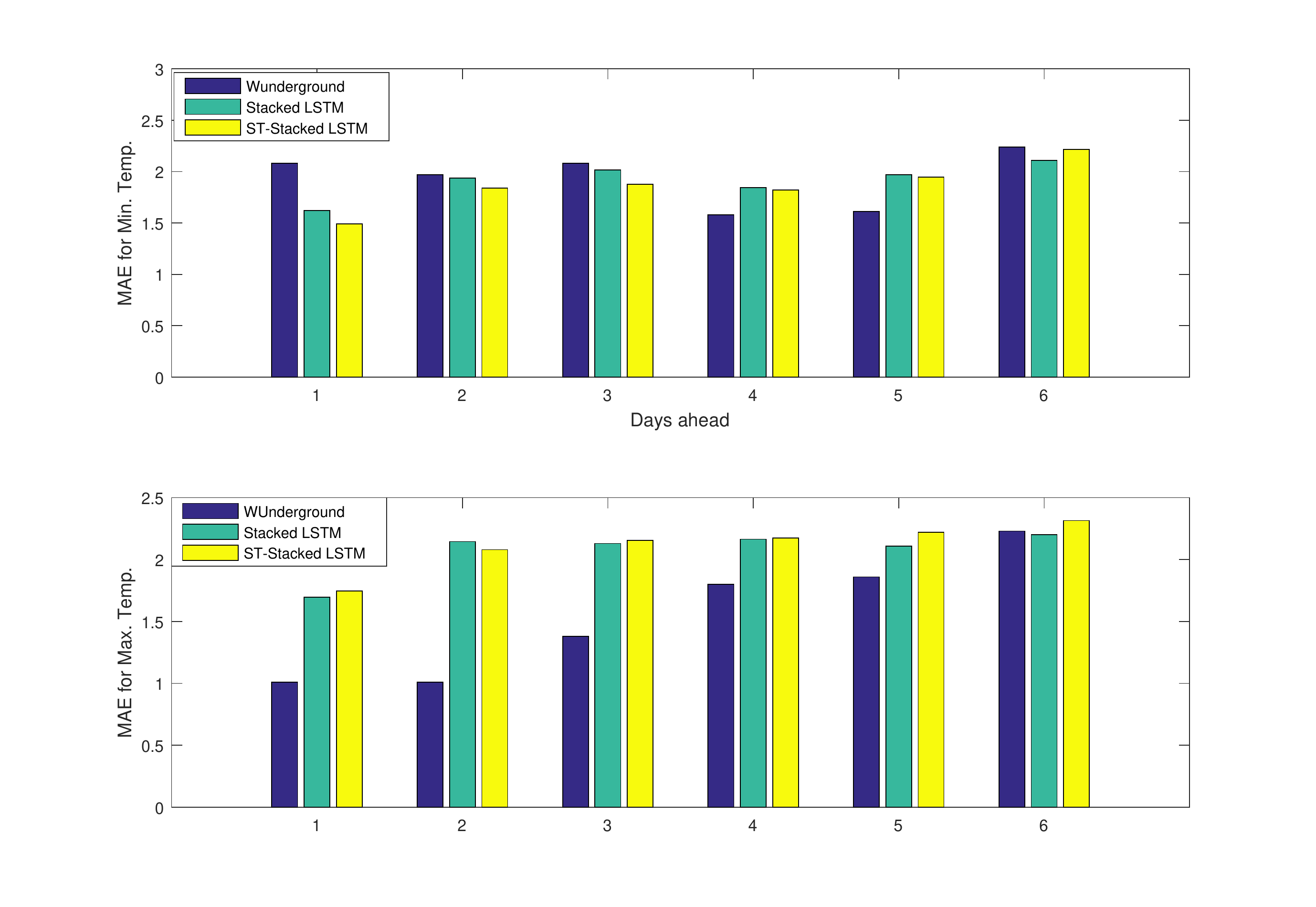}
		\captionof{figure}{Comparing MAE of min. and max. temperature prediction for Weather Underground, stacked LSTM, and Spatio-Temporal stacked LSTM (ST stacked LSTM) \label{ResAllST}}

   \end{multicols}

%
%
%


\vspace{-0.5cm}


%

\bibliographystyle{acm}
\bibliography{refs}
%
%
%

\end{document}